\documentclass[a4paper,12pt]{article}
\usepackage{fancyhdr,graphicx}
\begin{document}

\title{An Integrated Software-based Solution for Modular and Self-independent Networked Robot}

\author{I. Firmansyah, Z. Akbar, B. Hermanto and L.T. Handoko\\
Group for Theoretical and Computational Physics, Research Center for Physics\\ Indonesian Institute of Sciences (LIPI)\\
Kompleks Puspiptek Serpong, Tangerang 15310, Indonesia\\
firmansyah@teori.fisika.lipi.go.id, zaenal@teori.fisika.lipi.go.id}

\maketitle

\thispagestyle{fancy}
\fancyhead{}
\lhead{}
\cfoot{}
\rfoot{}
\renewcommand{\headrulewidth}{0pt}
\renewcommand{\footrulewidth}{0pt}

\begin{abstract}
An integrated software-based solution for a modular and self-independent networked robot is introduced. The wirelessly operatable robot has been developed mainly for autonomous monitoring works with full control over web. The integrated software solution covers three components : a) the digital signal processing unit for data retrieval and monitoring system; b) the externally executable codes for control system; and c) the web programming for interfacing the end-users with the robot. It is argued that this integrated software-based approach is crucial to realize a flexible, modular and low development cost mobile monitoring apparatus.\\

\end{abstract}

\section{Introduction}

During the last decades, automated systems embedded with remote robots are getting common in all aspects of human life. In some applications, it is motivated in most cases by the concern of safety as volcano observations and so forth. Furthermore, instead of radio-frequency based remote robots, in recent years the remote robot is evolutionary advancing to be networked robot. The technology migration is of course possible due to the advancements in internet technologies. Many groups have developed the so-called networked robots that is robotic systems controlled remotely over internet using TCP/IP protocol. Most of them fall into the category  to support human daily life, or to realize more interactive humanoids. One example is the WAX Project which is the second tele-operated internet robot at Ryerson Polytechnic Universities, the MAX Project \cite{max3}. The MAX Tele-Operated Dog has shown that  a tele-operated robot controlled from over web is quite reliable \cite{max1}\cite{max2}. On the other hand, the WAX puts together a procedure to change any robot into a tele-operated robot on the web \cite{wax}. Either MAX / WAX or MONEA are the microcontroller-based robots equipped with onboard computer, camera and  microphone with a main purpose to simulate telepresence. Originally these robots were intended and  simulated to support the handicapped persons. So the main issue is how to recognize the captured images or sounds and interpret them to be useful information for potential users.

Another kind of networked robot is the MONEA (Message-Oriented NEtworked-robot Architecture) which is an efficient development platform architecture for multifunctional robots \cite{monea}. The architecture embeds a Networked-Whiteboard Model for information sharing framework along with Message passing framework via P2P Virtual Network using Interest-Oriented Module Groups and Software Patterns to reduce complexity risks. It has actually been designed to fulfill three features : embodying the Meta-Architecture for Networked-Robots, supporting Bazaar-Style Development Model, and no need of heavy weight middleware. The MONEA-based middleware has been implemented to develop a dialog robot for exhibition.

On the other hand, there is also another usage of tele-operated mechanism to control and monitor simultaneously several robots over web \cite{amire}. The system provides a comprehensive platform  enabling the users to customize each robot independently.

In our present project of LIPI Networked Robot (LNR) we follow similar approach to develop a modular and self-independent wireless robot \cite{lnr1}. The difference is on the main objective to perform more serious tasks requiring telepresence for the reason of safety. For example : direct data retrieval in nuclear reactors, observation apparatus for volcanoes and so on. These kinds of purpose lead to completely different requirements. The robot should be able to acquire data in almost real-time basis, and then to process it at the robot's local system as well. Therefore no need for the end-users to install certain softwares in a terminal connecting to the robot. The users just connect their terminals over wireless network and then pointing the browser to the assigned address of robot to display the analyzed results, while at same time it is able to control and monitor the robot's movement, direction etc. Also, by definition there is no requirement on neither complex physical movement nor human interfaces.

The present paper is focused on describing the whole aspects of software-based solution in LNR. Because the detailed hardware components in LNR has been presented previously \cite{lnr2}\cite{lnr3}. Further, a real implementation in LNR as the prototype is discussed. Finally we summarize the results and mention future plans.

\section{The concept}

\begin{figure*}[t]
 \centering
 \includegraphics[width=16cm]{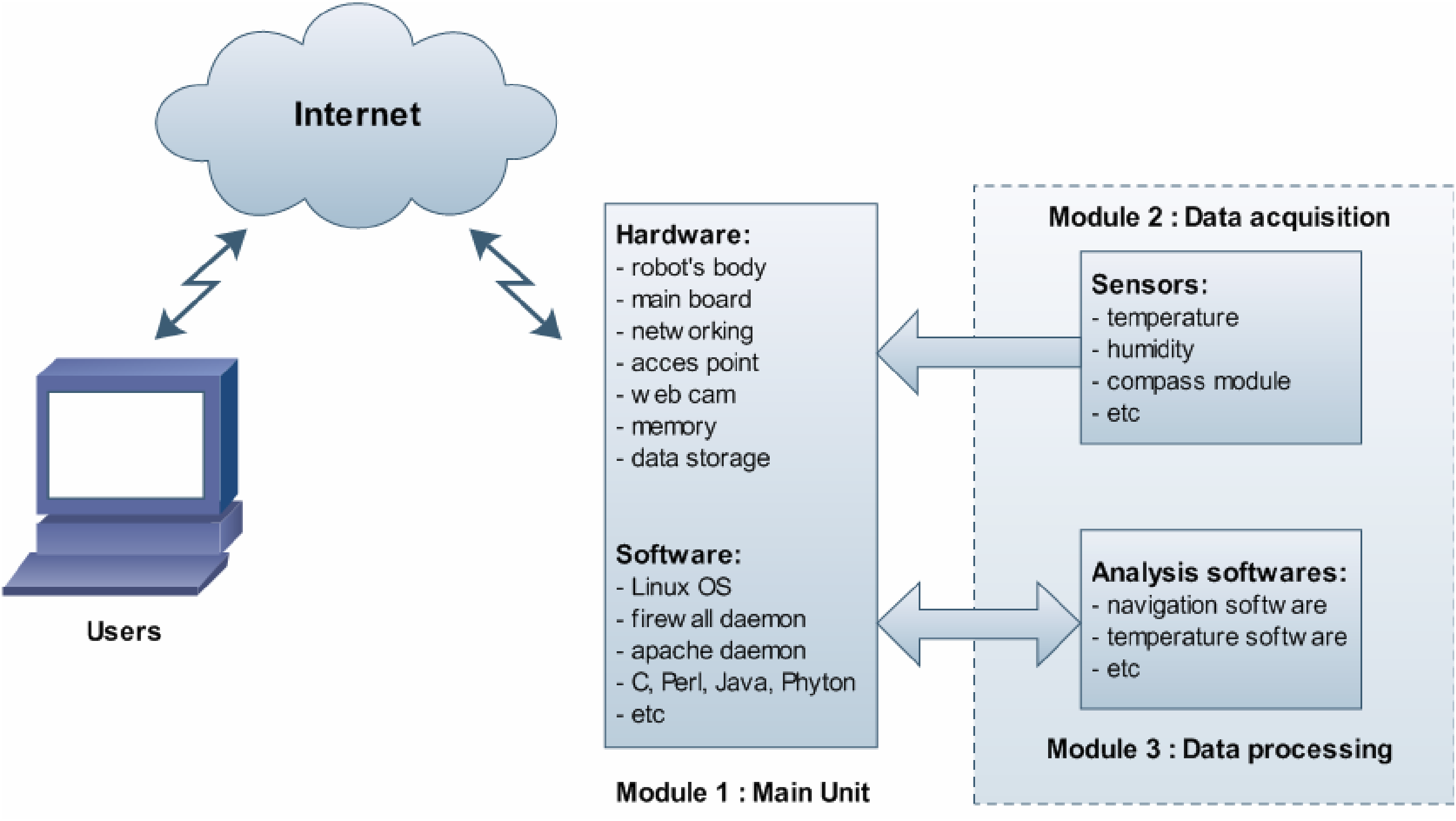}
 \caption{Three modules and its contents constructing the modular robot.}
 \label{fig:modul}
\end{figure*}

Here, let us again review briefly the concept to realize the features mentioned above. We have developed the robot to be as modular as possible to make it more adaptive to users' needs. This concept is depicted in Fig. \ref{fig:modul}, consisting of three modules \cite{lnr2},
\begin{enumerate}
\item Main unit. 
\item Data acquisition module.
\item Data processing module.
\end{enumerate}
The first module mainly involves hardwares of actuators, network and embedded PC. The second one contains combination of microcontrollers and the related softwares for signal processing, while the last one is completely a software-based solution. Of course, the main unit is also responsible for the underlying operating system including mini web server and so forth. In fact, the embedded data processing system enables the robot to be self-independent.

The last two modules are closely integrated each other in the sense that two modules should always belong to the same package. So, the system is practically divided into two independent units : the main unit and the Data Acquisition and Processing Systems (DAPS) unit for data acquisition and its processing modules. In principle, one might have single main unit with different packages of DAPS, or vice versa according to the needs. This characteristic would enable users to easily replace, for instance the set of censors and relevant software-based data processing modules with any available packages later on using the same main unit. Inversely, replacing the main unit with more appropriate one to a certain geographical site, but using the same DAPS package.

Now, the whole architecture is available for public as an open source \cite{opennr}. We expect that this approach could  encourage third parties to develop independently any relevant packages designed for some certain needs. Inversely, another ones who are interested more in hardware developments, might build alternative main unit with different type of mechanics for robot, but with same architecture for the rest to keep its compatibilities with existing DAPS packages. Some considerable packages are, for example, censors of hazard gas combined with software based chemical compound analyzer, vibration censor combined with software based seismograph, and so forth. This is the reason we call it as a generic robot which is adaptable to any DAPS. 

In contrary most  robots are usually constructed for single particular task. According to this concept, in order to realize high degree of freedom for both users and developers, and to keep full compatibilities in the future developments, let us list common features which should be fulfilled :
\begin{enumerate}
\item All aspects are fully controllable wirelessly over web through TCP/IP protocol.
\item Acquired data are stored and processed at the robot's local system independently from external apparatus.
\item The processed data can be retrieved and analyzed by users also over web such that no need for additional software installation at the user's terminal connecting to the robot.
\item The hardware-based components are replaced as much as possible with the software-based systems, even in the main unit. 
\end{enumerate}
According to the last point, in particular signal processing and filtering from the censors are performed using customized software rather than hardware as commonly done. This is important again to improve the flexibility and overall cost reduction. Hence, attaching different type of censors would require only different set of DAPS accordingly.

list here some advantages in deploying this approach for a mobile monitoring robot :
\begin{itemize}
\item Easy access for end-users regardless the operating system being used. 
\item No need for installing any additional softwares in terminal accessing the robot. 
\item Overall cost reduction, since most components are replaced with software based system.
\item High compatibility due to limited proprietary hardwares and also embedded softwares in the system. \item All software based components are developed using freely available open-source softwares. In our case we use Debian Linux for the operating system, Apache for the web-server and some GNU Public License development languages like GNU C, Java and Python.
\item Highly safe at any untouchable areas of human being, since all aspects of robot are remotely accessible and controllable.
\item Simple calibration since all signals are processed by softwares. This point is crucial regarding the main purpose of direct and continuous retrieval of  physical observables.
\end{itemize}

\section{Software-based solutions}

Now we are ready to introduce the software-based solutions for a modular and self-independent robot. Following the architecture previously mentioned, the software can also be divided into three parts. 

\subsection{Main unit software}

The software are embedded in the main unit, involving the operating system, web server, networking, some interpreters like Java and shell.

Beside some standard based system, networking and programming tools, the main unit software also contains some codes for the control system of actuators. The actuators might be the robot mechanics or some supplementing tools like camera and so on.

In some cases it could also involve a kind of monitoring system, that is processing the feedback from  actuators and reflecting it back to the control system to enable an intelligent robot. Needless to say, the software should process the navigation system as well, either using the image processing or dead-reckoning method.

Finally the main unit software is responsible for interfacing the whole system with the users. We have deployed a user-friendly and integrated web interface covers :
\begin{itemize}
\item Front module for for common information.
\item Administration module to maintain the whole parameters, basic settings, authenticate a new user or terminal, uploading or upgrading the data acquisition and processing software relevant for particular hardwares being used, etc.
\item User module to enable user access to control and monitor the main unit, to display or download the raw data or analyzed results.
\end{itemize}

\subsection{Data acquisition software}

This is mainly the digital signal processing (DSP) system \cite{stein}. Because, for the sake of improving the modularity, all acquired data from the attached censors through microcontrollers are not processed by the hardware-based filters as commonly done. Instead, the raw signals are sent directly to the main unit to be processed and filtered using the dedicated software.

This approach would definitely require developments on dedicated DSP's along with the user needs. The benefits are clear, i.e. :
\begin{itemize}
\item It enables unlimited combination on processing the signals by just adjusting the relevant parameters in the DSP system.
\item Periodical adjustment or calibration can be done easily whenever it is needed. This would improve the accuracy of data taken by censors etc.
\item On the other hand, it simplifies the data acquisition hardwares that further improves its flexibilities and modularities. Because, for instance we can in principle attach any censors to the main unit and modify the data acquisition software accordingly. 
\end{itemize}

\subsection{Data processing software}

The most unique component is the data processing software. This component plays an important role to get rid of any software installation at the user's terminal connecting to the robot. The user just runs the browser, points it to the assigned address and that's all.

As its nature, the data processing software should be a dedicated software for a particular purpose relevant with the acquirable data taken from a set of data acquisition hardwares.

We should remark here that the web interface in the main unit has a capability to install a new or upgrade newer version of DAPS software over web. 

\section{A prototype : LIPI Networked Robot}

For the sake of completeness, let us discuss the case of LNR that is accessible on the net for public \cite{lnr1}. Since we put the priorities on the DAPS unit rather than the main unit, LNR adopts a simple actuators to enable it to move in a limited space.

To be more specific, we have designed a DAPS package to observe hazard gas in a certain area and to measure the surrounding environment. The whole system then consists of,
\begin{itemize}
\item Main unit :\\
	Containing main processor (mini PC etc), storage media (mini hard-disk), access point, power supply, battery and all mechanical components. Of course, it also includes the underlying operating system,  integrated web interface,  hardware control and monitoring systems  and storing system for all  data.
\item Data acquisition module :\\
	A set of censors ($CO$ gas, temperature, humidity, $NO$ gas, smoke) and small camera. The softwares in this  module are also responsible for filtering the raw signal from censors. 
\item Data processing module :\\
	It covers all add-on softwares to process, store and analyze the acquired data and display the results accordingly. The system provides a real-time warning system regarding the air quality in a certain area.
\end{itemize}

For communication between the main unit and the control and data acquisition hardwares we use the  parallel port for both reading the signals from and sending some commands to microcontroller. As a result, the microcontroller decodes each command into an associated task such as driving the actuators (in the case of LNR are two DC motors), and also retrieving the raw signals from attached censors. 

In this paper, however we are not going to discuss the microcontrollers in the DAPS unit, nor the communication hardwares itself. Because those details are out of coverage of the paper, and have already been discussed in our previous works \cite{lnr2}\cite{lnr3}. Instead, we focus on describing briefly our  concept to deal with the wheels as actuators and the associated dead reckoning to track the current position easily with very limited hardwares and mostly relying  on software based systems. Instead we discuss in more detail the algorithm we use to control and to obtain some feedback from the actuators. Again, the detail about the actuators in LNR should refer to \cite{lnr3}.

The actuators in LNR enables movements in any directions by deploying two independent and one free wheels \cite{braunl}\cite{mccomb}. This architecture is to avoid unnecessary complexities. The mechanical components consists of three wheels, one is freely rotatable while the other two are fixed and driven by two independent DC motors. Using these motors we are able to control the robot movement in four directions such as turning left and right, moving forward and backward by controlling each DC motors independently. Turning left (right) is realized by stopping the left (right) motor and starting the right (left) motor respectively. Moving forward is simply done by starting both motors simultaneously.

The next problem is how to recognize the current position at almost real-time basis. Instead of using the images captured by camera which would require complicated and resource wasting image processing, we make use of the compass censor and the wheel rotation counters. Again, we remind that in LNR the main processor should be prioritized to retrieve, store and analyze the data.  Although a small camera is attached on the robot, it is only intended to get the actual visualization around the robot.  Therefore, the problem is turned into how to track the paths and recognizing the current position in a sightless condition. 

The compass censor provides information of the actual angle against the North-South pole, while counting each wheel rotation yields the point-to-point distance. This mechanism is later on visualized on the web in real-time basis as a virtual compass and the footprints of wheels from one point to another. This compass module has been specifically designed for a usage in robots as add-on navigation. It generates a unique number in byte from 0 to 255 to represent the angles within $0^\mathrm{o}\sim360^\mathrm{o}$. The internal microcontroller in the compass module then converts the signals from magnetic field censor into serial I2C data format. Using the compass module, the angles information is retrieved through I2C communication protocol. This allows us to read one or two bytes for 8-bit or 16-bit registers respectively. 

In order to recognize a real path length, we simply count the wheels rotation whenever the wheels start moving and stopping. The wheel rotation is measured by using opt coupler. The opt coupler detects a black and white colors on the wheel and produces digital signals (1 and 0). The data is acquired by microcontroller and stored in the main unit storage media. In LNR each wheel is divided into eight areas which means a complete rotation occurs after eight counts, however we can divide it as much as possible to improve the accuracy. All information of the angles and the path lengths are recorded path by path. Finally we borrow the dead reckoning method which is very powerful to calculate the position relying on a previously determined position \cite{dr}. To implement such algorithms on a robot, we have constructed a mathematical prescription to calculate the geometrical distances of each path obtained from each wheels counter, and its relative angles obtained from the compass censor. The final formula is embedded into the software belonging to that main unit. The formula provides an exact calculation for each intermediate path, the total running-distance over the paths and the real distance between the initial and end points. Nevertheless, due to the limited space the detail algorithm and formulae will be given elsewhere.

Below is a short and simple example on the algorithm to control the motors in LNRusing C language. 
\begin{footnotesize}
\begin{verbatim}
#include <stdio.h>
#include <stdlib.h>
#include <unistd.h>
#include <sys/io.h>

#define base 0x378     
#define status base+1  
#define control base+2 

main(int argc, char **argv)
{

 int input=0;

 if (ioperm(base,1,1))
  fprintf(stderr, "Error: port %x\n", base), 
     exit(1);
 if (ioperm(status,1,1))
  fprintf(stderr, "Error: port %x\n", status), 
     exit(1);
 if (ioperm(control,1,1))
  fprintf(stderr, "Error: port %x\n", control), 
     exit(1);

    outb(0x01, base);		//to set the robot move forward
    usleep(5000);

\end{verbatim}
\end{footnotesize}

The value of data sent to the microcontrollers depends on the initial setting. In this case we use the value of \texttt{0x01}. The algorithm written above is enough to send the command to the robot since there is no need for feedback so that it will move the robot in forward direction. In contrary, for the compass censor one needs more as below to read the data from the compass censor. Algorithm below shows how to read the data using control address of parallel port in nibble mode with the help of IC 74LS151.

\begin{footnotesize}
\begin{verbatim}
outb(0x09, base);		//command to read compass
usleep(5000);

outb(inb(control)|0x01,control);
input=(inb(status)&0xF0);
usleep(100);		
input=input>>4;

outb(inb(control) & 0xFE,control);	// strobe = 1
input=input|(inb(status)&0xF0);
usleep(100);	
input=(input^0x88);
usleep(100);
printf("compas : %d \n",input);
\end{verbatim}
\end{footnotesize}

\section{Summary and discussion}

We have introduced an integrated software-based solution for modular networked robot which is fully controllable and accessible over network. As a typical implementation of the concept, we have developed the LNR. We have shown that the software-based solution is crucial to enable the required features. Also it improves significantly its flexibility to various purposes. We argue that regarding its main objective as a monitoring apparatus, LNR is quite efficient and has good total cost-performance due to its modularity and dominant software based solutions. 

However, the system has complicated aspects and still requires further developments as,
like :
\begin{itemize}
\item More examples of DAPS packages fit certain purposes. 
\item More complicated robot's actuators and the relevant algorithms for its software-based control and monitoring systems.
\item Further development of main unit and its mechanical components to enable more advanced and smooth movements. Rather than full hardware-based approach, this will be done by utilizing as much as possible integrated software-based algorithms.
\item Automatic and software based calibration systems to keep the accuracy of actuators. For instance  in the current LNR, synchronizing the rotation speed of left and right wheels, reseting the initial angle of camera and so on.
\item Lastly, we would like to announce that the architecture and all related softwares of LNR will be open for public under GNU Public License once we consider the system is ready for further development by open-source communities around the world \cite{opennr}.
\end{itemize}

\section*{Acknowledgment}

The work is financially supported by the Riset Kompetitif LIPI in fiscal year 2008.

\end{document}